%% file: acl_latex.tex
\pdfoutput=1

\documentclass[11pt]{article}

\usepackage[final]{acl}

\usepackage{times}
\usepackage{latexsym}

\usepackage{booktabs}
\usepackage{multirow} 
\usepackage{booktabs}
\usepackage{multirow}
\usepackage{geometry}
\usepackage{adjustbox}
\usepackage{enumitem}

\usepackage[T1]{fontenc}

\usepackage[utf8]{inputenc}

\usepackage{microtype}

\usepackage{inconsolata}

\usepackage{graphicx}

\title{Do Large Language Models Know Conflict? Investigating Parametric vs. Non-Parametric Knowledge of LLMs for Conflict Forecasting}
\author{
  \textbf{Apollinaire Poli Nemkova\textsuperscript{1}},
  \textbf{Sarath Chandra Lingareddy\textsuperscript{1}},
  \textbf{Sagnik Ray Choudhury\textsuperscript{1}},
  \textbf{Mark V. Albert\textsuperscript{1}}
\\
\\
  \textsuperscript{1}University of North Texas, USA
\\
  \small{
    \textbf{Correspondence:} \href{mailto:poli.nemkova@dunt.edu}{poli.nemkova@unt.edu}, 
    \href{mailto:sarathchandralingareddy@my.unt.edu}{sarathchandralingareddy@my.unt.edu}, 
    \href{mailto:sagnik.raychoudhury@unt.edu}{sagnik.raychoudhury@unt.edu},
    \href{mailto:mark.albert@unt.edu}{mark.albert@unt.edu}
  }
}

\begin{document}
\maketitle

\begin{abstract}
Large Language Models (LLMs) have shown impressive performance across natural language tasks, but their ability to forecast violent conflict remains underexplored. We investigate whether LLMs possess meaningful parametric knowledge—encoded in their pretrained weights—to predict conflict escalation and fatalities without external data. This is critical for early warning systems, humanitarian planning, and policy-making.
We compare this parametric knowledge with non-parametric capabilities, where LLMs access structured and unstructured context from conflict datasets (e.g., ACLED, GDELT) and recent news reports via Retrieval-Augmented Generation (RAG). Incorporating external information could enhance model performance by providing up-to-date context otherwise missing from pretrained weights.
Our two-part evaluation framework spans 2020–2024 across conflict-prone regions in the Horn of Africa and the Middle East. In the parametric setting, LLMs predict conflict trends and fatalities relying only on pretrained knowledge. In the non-parametric setting, models receive summaries of recent conflict events, indicators, and geopolitical developments.
We compare predicted conflict trend labels (e.g., Escalate, Stable Conflict, De-escalate, Peace) and fatalities against historical data. Our findings highlight the strengths and limitations of LLMs for conflict forecasting and the benefits of augmenting them with structured external knowledge.

\end{abstract}

\section{Introduction}
\input{intro}

\section{Literature Review}
\input{litrev}

\section{Method}
\input{method}

\section{Results}
\input{results}

\section{Discussion}
\input{discussion}

\section{Conclusion}
\input{conclusion}

\section*{Limitations}
This study presents a first step toward evaluating LLMs for conflict forecasting, but several limitations remain. Our test sets are relatively small at the country level (59 monthly points), though expanded across regions for broader coverage. Class definitions rely on slope thresholds, which may oversimplify complex conflict dynamics. Additionally, the quality of RAG outputs varies depending on retrieved content and summarization, sometimes leading to imprecise or hallucinated estimates. Finally, we evaluate only general-purpose LLMs in zero-shot settings; fine-tuned models or multilingual pipelines could yield stronger results.

Key limitations:
\begin{itemize}[noitemsep, topsep=0pt]
    \item Small per-country test size; regional heterogeneity introduces noise;
    \item Skewed data: instances of Stable/Conflict and Peace are more common than Escalate or De-escalate;
    \item Threshold-based labeling may not capture nuanced escalation patterns;
    \item RAG summaries can introduce ambiguous or noisy context;
    \item No fine-tuning or few-shot adaptation used;
    \item English-only evaluation; multilingual sources not yet explored.
\end{itemize}

\section*{Acknowledgments}

The authors appreciate the advice and guidance of Sun-joo Lee.

\bibliography{acl_latex}

\appendix



\end{document}

%% file: intro.tex
Forecasting violent conflict is a critical yet persistently challenging task. Humanitarian organizations, early warning systems, and policy actors rely on timely conflict predictions to guide resource allocation, diplomatic engagement, and civilian protection strategies. Traditionally, these forecasts are derived from structured statistical models, domain expert input, and region-specific knowledge, often leveraging structured event data such as ACLED or GDELT. However, these approaches require manual feature engineering, are brittle to new contexts, and may lack the capacity to generalize across diverse geopolitical settings.

With the rapid progress of large language models (LLMs), a new question emerges: can LLMs forecast conflict directly from their parametric knowledge—the information implicitly encoded in their pretrained weights? Given that many LLMs are trained on massive corpora of news articles, Wikipedia, and geopolitical documents, it is plausible that they have internalized latent patterns related to conflict escalation, resolution, and regional dynamics. The question of event forecasting using LLMs has been approached from a similar perspective by several authors \cite{Kauf2022Event,Ye2024MIRAI}.

In this work, we explore the feasibility of using LLMs to forecast conflict dynamics across two experimental settings:

- Parametric forecasting, where LLMs generate predictions using only their internal knowledge, without any access to recent or region-specific data.

- Retrieval-Augmented Generation (RAG), where models are supplied with structured indicators and a news-derived summary covering the previous three months, and are asked to forecast the following month.

We frame conflict forecasting as a dual task: (1) classifying the expected direction of violence—escalation, de-escalation, stable conflict, or peace; and (2) predicting the estimated number of fatalities in the forecast window.

Our experiments span multiple conflict-prone regions—including the Horn of Africa and the Middle East —over a five-year period (2020–2024). We evaluate several LLMs (GPT-4\footnote{https://openai.com/index/hello-gpt-4o/}, LLaMA-2\footnote{https://huggingface.co/meta-llama/Llama-2-7b}) using both zero-shot and RAG-based approaches. Results show that zero-shot LLMs capture broad conflict patterns (especially peaceful or stable dynamics), but struggle with finer-grained trends and numeric estimation. RAG-based augmentation, particularly in the monthly setting, improves macro-level metrics and fatalities regression. 

Interestingly, we find that while GPT-4 generally outperforms LLaMA-2 across most metrics, the open-source model shows promising capabilities in specific contexts, particularly in categorical predictions for certain regions. This suggests potential for open-source LLMs in humanitarian forecasting applications, especially in resource-constrained environments where proprietary model access may be limited. However, significant performance gaps remain that will require further research to address.

Our contributions are threefold:
(1) We develop a structured framework for evaluating LLMs in conflict forecasting tasks.
(2) We compare parametric and RAG-based capabilities across multiple models and regional contexts.
(3) We provide qualitative and quantitative insights into the limitations and strengths of current LLMs in high-stakes forecasting tasks, offering guidance for future work in AI-assisted early warning systems.

%% file: litrev.tex
Forecasting political violence and armed conflict has long been a priority for scholars and practitioners of early warning systems. Traditional approaches often rely on statistical models using structured conflict data, such as ACLED\footnote{https://acleddata.com/} or UCDP-GED\footnote{https://ucdp.uu.se/downloads/}, combined with socio-political and economic indicators. Recent advances in machine learning have enriched this landscape by enabling models to integrate large-scale unstructured data. For instance, the ViEWS project \cite{hegre2019views} demonstrated that political violence could be forecasted with subnational granularity using an ensemble of ML classifiers trained on conflict and governance data. Similarly, \citeauthor{attina2022forecasting} introduced the Dynamic Elastic Net (DynENet), which leverages over 700 variables—including event data from GDELT\footnote{https://www.gdeltproject.org/} and ACLED—to capture changes in conflict fatalities across countries. Other authors would utilize protest data for the task of conflict prediction \cite{Rød2023Predicting} or utilize transformer architecture for human rights violation detection \cite{nemkova2023detecting} as a part of conflict monitoring.

Text-based modeling has emerged as a particularly powerful tool in this space. \citeauthor{mueller2018reading} (2018) pioneered a method that forecasts conflict onset by extracting topic distributions from millions of newspaper articles. These topic features proved predictive of violent escalation even in countries with no prior conflict, offering early warning capability unattainable through historical trends alone. Their openly released dataset of monthly country- and district-level forecasts \cite{mueller2024introducing} has become a valuable benchmark for evaluating risk-aware conflict interventions.

While traditional ML approaches require task-specific engineering, Large Language Models (LLMs) offer a flexible alternative by generalizing across domains. Recent work has shown that LLMs, such as GPT-3 and LLaMA-2, are capable of performing temporal reasoning and event forecasting in zero-shot settings \cite{chang2024comprehensive, yu2306temporal, wang2024news}. The RAF framework \cite{tire2024retrieval} shows that foundation models can generate accurate continuations of time series without dedicated fine-tuning. However, LLMs trained purely on static corpora tend to struggle with up-to-date, domain-specific predictions. As such, their utility in high-stakes tasks like conflict forecasting remains underexplored.

Retrieval-Augmented Generation (RAG) offers one solution to the limitations of static parametric models by injecting fresh, context-rich information into the generation pipeline. \citeauthor{yang2025timerag}  introduced TimeRAG, which retrieves relevant time series segments to boost LLM forecasting performance. Similarly, TS-RAG \cite{ning2025ts} combines time series memory with large-scale retrieval to improve accuracy in zero-shot prediction tasks. These RAG-based frameworks are especially valuable in settings where recent developments can rapidly shift forecast dynamics. In the conflict domain specifically, \citeauthor{wood2024conflict} proposed Conflict-RAG, a multilingual RAG pipeline that enhances LLMs’ ability to understand current conflict events by retrieving regional news sources, including Arabic-language articles. Their approach improves situational awareness in rapidly evolving crises and demonstrates the promise of RAG-augmented LLMs for early warning systems.

Additionally, the phenomenon of knowledge conflict in LLMs \cite{xu2024knowledge} — where parametric knowledge encoded during pretraining contradicts non-parametric information retrieved at inference time—raises important questions for conflict forecasting. This makes it especially valuable to experimentally compare LLM performance in parametric-only versus retrieval-augmented settings, to evaluate not only overall accuracy but also the models’ ability to reconcile conflicting information \cite{Mallen2022When, Wang2023Self-Knowledge, jiang2024towards} . Recent work by \citeauthor{wu2024clasheval} demonstrates that some LLMs exhibit a stronger bias toward their parametric knowledge, even when provided with relevant external context via RAG, highlighting the need for model-specific evaluation in high-stakes domains. 

Together, these lines of research reveal four important trends: (1) media-derived features can provide strong signals for early conflict prediction, particularly for first-onset cases; (2) LLMs can offer flexible, language-aware forecasting capabilities, but require external knowledge to achieve reliability; (3) RAG is emerging as a bridge between deep language models and dynamic real-world data, with growing relevance to humanitarian and geopolitical forecasting tasks; and (4) the strength of context knowledge is limited when it conflicts with parametric contextual evidence.

While prior work has explored conflict forecasting using structured models, media-based signals, and more recently, large language models, limited attention has been given to systematically comparing parametric and non-parametric LLM reasoning for this task \cite{Zhang2024Leveraging, Zhang2024Evaluating}. In particular, the extent to which LLMs encode useful forecasting priors in their pretrained weights versus their ability to synthesize external geopolitical context remains underexplored. Our work addresses this gap by evaluating LLMs across three settings: (1) zero-shot forecasting using only parametric knowledge; (2) retrieval-augmented forecasting based on recent news summaries and structured indicators over a 3-month window. We benchmark multiple LLMs (GPT-4, LLaMA-2) across regions in the Horn of Africa and the Middle East  from 2020–2024, analyzing both trend classification and fatality prediction. This study provides a practical evaluation framework and new empirical insights into how well LLMs can support real-world early warning efforts—either on their own or when augmented with timely conflict data.

%% file: method.tex
We conduct a structured set of experiments to evaluate large language models (LLMs) in forecasting violent conflict. Specifically, we assess both parametric reasoning (pretrained knowledge alone) and non-parametric reasoning (via retrieval-augmented generation, or RAG). All experiments are conducted using two LLMs: GPT-4 (via OpenAI API \footnote{https://platform.openai.com/docs/overview} ) and LLaMA-2-13B-chat (via Hugging Face Transformers\footnote{https://huggingface.co/docs/transformers/en/index}).

\input{res_table_gpt}

\subsection{Data Sources}
Our dataset integrates both structured conflict data and open-source news articles. We use the Global Database of Events, Language, and Tone (GDELT) \cite{leetaru2013gdelt} as the primary source of unstructured information. GDELT provides metadata on global news coverage, including article URLs, actor country codes, event tone scores, and Goldstein Scale values (a proxy for cooperation–conflict dynamics). We queried GDELT for multiple regions—including the Horn of Africa and the Middle East —between 2020 and 2024, and scraped article content using the provided URLs.

Structured conflict outcome data comes from the Armed Conflict Location and Event Data Project (ACLED), which records real-world conflict incidents, including the number of fatalities. Fatality data is used to construct ground truth for both classification and regression tasks. In Experiment 2, we also use average article tone and average Goldstein scores \cite{mcclelland1984world} from GDELT as numerical features to contextualize retrieved summaries.

\subsection{\textit{Experiment 1: Parametric Forecasting}}
In the first setting, we test whether LLMs can forecast conflict using only their internal parametric knowledge. The model is prompted to generate a one-month forecast for a given country, based on recent conflict and social dynamics inferred from its pretrained corpus. It must classify the conflict trend using one of four predefined labels: Escalate, De-escalate, Peace/No Conflict, Stable Conflict. (We follow the labeling system as  temporal states utilized in \cite{croicu2025newswire}).

Additionally, the model is asked to estimate the expected number of fatalities during that period, expressed either as a numerical range or specific count.

\subsubsection{\textit{Experiment 2: Retrieval-Augmented Forecasting (RAG)}}
In the second setting, we assess whether access to structured and unstructured context improves forecasting accuracy. Using a RAG pipeline, the LLM is supplied with retrieved summaries and metadata covering the previous three months and is asked to forecast the next month using the same label set and fatality estimate. The retrieval step is performed using the FAISS library (Facebook AI Similarity Search)\footnote{https://ai.meta.com/tools/faiss/}, while summarization is conducted using GPT-3.5. The provided context includes:

\begin{itemize}
    \item A summary of the most relevant news excerpts (retrieved and summarized via semantic search);
    \item The average tone of retrieved articles (from GDELT);
    \item The average Goldstein Scale score (from GDELT);
    \item Weekly fatality counts over the past 12 weeks (from ACLED).
\end{itemize}

\subsection{Evaluation Framework}
We evaluate model outputs across three prediction types:

\begin{enumerate}
    \item Categorical label prediction (e.g., Escalate)
    \item Fatality range prediction (binned using quantiles)
    \item Label derived from predicted fatalities, matched to true class thresholds (e.g., 400 fatalities corresponds to Stable Conflict)
\end{enumerate}

Metrics include accuracy, precision, recall, and F1-score, reported in micro, macro, and weighted forms for classification tasks. For regression, we report Mean Absolute Error (MAE).

\subsection{Implementation Details}
All experiments are run on Google Colab Pro+ \footnote{https://colab.research.google.com/} using NVIDIA A100 GPUs. GPT-4 queries are queried via OpenAI’s GPT-4 API endpoint with temperature set to 0.2. The total OpenAI API cost for this study was \$150. Prompt templates  and all source code is available on the author’s GitHub repository\footnote{https://github.com/anonymous-author}.

\input{res_table_llama}

%% file: res_table_gpt.tex
\begin{table*}[h]
\centering
\caption{GPT-4 Evaluation Metrics – Experiment 1 (Parametric) vs. Experiment 2 (RAG)}
\begin{adjustbox}{max width=\textwidth}
\begin{tabular}{llccccc}
\toprule
\textbf{Experiment} & \textbf{Metric} & \textbf{Ethiopia} & \textbf{Sudan} & \textbf{Somalia} & \textbf{Israel} & \textbf{Iran} \\
\midrule

\multirow{5}{*}{Exp 1: Class (Categorical)} 
& Accuracy            & 0.2712 & \textbf{0.7458} & 0.1695 & 0.4098 & 0.4667 \\
& Precision (macro)   & 0.2496 & 0.1864 & \textbf{0.3092} & 0.1096 & 0.1750 \\
& Recall (macro)      & \textbf{0.4690} & 0.2500 & 0.3118 & 0.2232 & 0.1556 \\
& F1 (macro)          & 0.2091 & \textbf{0.2136} & 0.1171 & 0.1471 & 0.1647 \\
& F1 (weighted)       & 0.2728 & \textbf{0.6372} & 0.1682 & 0.2700 & 0.4941 \\

\multirow{5}{*}{Exp 2: Class (Categorical)} 
& Accuracy            & 0.2712 & \textbf{0.7288} & 0.1356 & 0.3770 & 0.3385 \\
& Precision (macro)   & 0.2496 & \textbf{0.2995} & 0.2112 & 0.2738 & 0.2741 \\
& Recall (macro)      & \textbf{0.4690} & 0.3773 & 0.1505 & 0.3736 & 0.2257 \\
& F1 (macro)          & 0.2091 & \textbf{0.3333} & 0.0964 & 0.2710 & 0.1980 \\
& F1 (weighted)       & 0.2728 & \textbf{0.6638} & 0.1693 & 0.3781 & 0.4285 \\

\midrule

\multirow{5}{*}{Exp 1: Class (From Fatalities)} 
& Accuracy            & 0.7288 & 0.7458 & \textbf{0.8305} & 0.4590 & 0.7500 \\
& Precision (macro)   & 0.1822 & 0.1864 & \textbf{0.2768} & 0.1148 & 0.1875 \\
& Recall (macro)      & 0.2500 & 0.2500 & \textbf{0.3333} & 0.2500 & 0.2500 \\
& F1 (macro)          & 0.2108 & 0.2136 & \textbf{0.3025} & 0.1573 & 0.2143 \\
& F1 (weighted)       & 0.6145 & 0.6372 & \textbf{0.7536} & 0.2888 & 0.6429 \\

\multirow{5}{*}{Exp 2: Class (From Fatalities)} 
& Accuracy            & 0.7288 & 0.7288 & \textbf{0.8305} & 0.4590 & 0.5231 \\
& Precision (macro)   & 0.1822 & \textbf{0.2995} & 0.2188 & 0.2524 & 0.2470 \\
& Recall (macro)      & 0.2500 & 0.3773 & 0.2500 & \textbf{0.4643} & 0.4222 \\
& F1 (macro)          & 0.2108 & \textbf{0.3333} & 0.2333 & 0.3252 & 0.2782 \\
& F1 (weighted)       & 0.6145 & \textbf{0.6638} & \textbf{0.7751} & 0.3440 & 0.5118 \\

\midrule

\multirow{1}{*}{Exp 1: MAE} 
& Fatalities MAE      & 446.59 & 512.78 & 326.41 & 218.48 & \textbf{176.25} \\

\multirow{1}{*}{Exp 2: MAE} 
& Fatalities MAE      & 446.59 & 517.56 & 362.51 & \textbf{93.54} & 96.80 \\

\midrule

\multirow{5}{*}{Exp 1: Binned Regression} 
& Accuracy            & 0.2542 & 0.2203 & \textbf{0.3220} & 0.0000 & 0.1833 \\
& Precision (macro)   & \textbf{0.1918} & 0.1214 & 0.1671 & 0.0000 & 0.0458 \\
& Precision (weighted)& \textbf{0.2081} & 0.1235 & 0.2062 & 0.0000 & 0.0336 \\
& Recall (macro)      & 0.2417 & 0.2167 & \textbf{0.2929} & 0.0000 & 0.2500 \\
& F1 (weighted)       & \textbf{0.2225} & 0.1464 & 0.2203 & 0.0000 & 0.0568 \\

\multirow{5}{*}{Exp 2: Binned Regression} 
& Accuracy            & 0.2542 & 0.3390 & 0.2373 & \textbf{0.3443} & 0.3385 \\
& Precision (macro)   & 0.1918 & \textbf{0.4404} & 0.2542 & 0.3166 & 0.3036 \\
& Precision (weighted)& 0.2081 & 0.4425 & 0.1895 & \textbf{0.5918} & 0.4033 \\
& Recall (macro)      & 0.2417 & 0.3429 & 0.2577 & 0.2072 & \textbf{0.3221} \\
& F1 (weighted)       & 0.2225 & \textbf{0.3483} & 0.1671 & 0.4280 & 0.3337 \\

\bottomrule
\end{tabular}
\end{adjustbox}
\label{table1}
\end{table*}

%% file: res_table_llama.tex
\begin{table*}[h]
\centering
\caption{LLama Evaluation Metrics – Experiment 1 (Parametric) vs. Experiment 2 (RAG)}
\begin{adjustbox}{max width=\textwidth}
\begin{tabular}{llccccc}
\toprule
\textbf{Experiment} & \textbf{Metric} & \textbf{Ethiopia} & \textbf{Sudan} & \textbf{Somalia} & \textbf{Israel} & \textbf{Iran} \\
\midrule

\multirow{5}{*}{Exp 1: Class (Categorical)} 
& Accuracy            & 0.0714 & \textbf{0.3898} & 0.0536 & 0.3770 & 0.0536 \\
& Precision (macro)   & 0.0374 & 0.1878 & 0.0144 & \textbf{0.2738} & 0.0153 \\
& Recall (macro)      & 0.3333 & 0.1667 & 0.2500 & \textbf{0.3736} & 0.2500 \\
& F1 (macro)          & 0.0670 & 0.1633 & 0.0273 & \textbf{0.2710} & 0.0288 \\
& F1 (weighted)       & 0.0144 & \textbf{0.4443} & 0.0058 & 0.3781 & 0.0062 \\

\multirow{5}{*}{Exp 2: Class (Categorical)} 
& Accuracy            & 0.0000 & 0.2542 & 0.4107 & \textbf{0.5439} & 0.0526 \\
& Precision (macro)   & 0.2083 & 0.2180 & 0.2054 & \textbf{0.3576} & 0.0150 \\
& Recall (macro)      & 0.2229 & 0.1655 & 0.1173 & \textbf{0.4018} & 0.2500 \\
& F1 (macro)          & 0.1933 & 0.1476 & 0.1494 & \textbf{0.3411} & 0.0283 \\
& F1 (weighted)       & 0.4732 & 0.3328 & \textbf{0.5227} & 0.5236 & 0.0060 \\

\midrule

\multirow{5}{*}{Exp 1: Class (From Fatalities)} 
& Accuracy            & 0.4821 & 0.7288 & 0.8393 & 0.4590 & 0.7321 \\
& Precision (macro)   & 0.2148 & 0.1853 & 0.2217 & \textbf{0.2524} & 0.1830 \\
& Recall (macro)      & 0.1869 & 0.2443 & 0.2398 & \textbf{0.4643} & 0.2500 \\
& F1 (macro)          & 0.1949 & 0.2108 & 0.2304 & \textbf{0.3252} & 0.2113 \\
& F1 (weighted)       & 0.5373 & 0.6288 & \textbf{0.8064} & 0.3440 & 0.6189 \\

\multirow{5}{*}{Exp 2: Class (From Fatalities)} 
& Accuracy            & 0.7288 & 0.6949 & \textbf{0.8750} & 0.4912 & 0.7368 \\
& Precision (macro)   & \textbf{0.3114} & 0.2708 & 0.2917 & 0.1228 & 0.1842 \\
& Recall (macro)      & \textbf{0.4050} & 0.3216 & 0.3333 & 0.2500 & 0.2500 \\
& F1 (macro)          & \textbf{0.3520} & 0.2940 & 0.3111 & 0.1647 & 0.2121 \\
& F1 (weighted)       & 0.6389 & 0.6367 & \textbf{0.8167} & 0.3236 & 0.6252 \\

\midrule

\multirow{1}{*}{Exp 1: MAE} 
& Fatalities MAE      & 624.66 & 575.25 & 344.91 & \textbf{93.54} & 370.80 \\

\multirow{1}{*}{Exp 2: MAE} 
& Fatalities MAE      & \textbf{572.49} & 649.92 & 369.63 & 229.23 & 376.84 \\

\midrule

\multirow{5}{*}{Exp 1: Binned Regression} 
& Accuracy            & 0.2679 & 0.2203 & 0.2500 & \textbf{0.3443} & 0.1786 \\
& Precision (macro)   & 0.1494 & 0.1533 & \textbf{0.3671} & 0.3166 & 0.0446 \\
& Precision (weighted)& 0.1434 & 0.2302 & 0.3532 & \textbf{0.5918} & 0.0319 \\
& Recall (macro)      & \textbf{0.2692} & 0.1892 & 0.3036 & 0.2072 & 0.2500 \\
& F1 (weighted)       & 0.1335 & 0.1919 & 0.1744 & \textbf{0.4280} & 0.0541 \\

\multirow{5}{*}{Exp 2: Binned Regression} 
& Accuracy            & 0.2203 & 0.2542 & \textbf{0.3214} & 0.0000 & 0.1754 \\
& Precision (macro)   & 0.1271 & 0.1834 & \textbf{0.4450} & 0.0000 & 0.0439 \\
& Precision (weighted)& 0.1547 & 0.2417 & \textbf{0.5075} & 0.0000 & 0.0308 \\
& Recall (macro)      & 0.2042 & 0.2237 & \textbf{0.3155} & 0.0000 & 0.2500 \\
& F1 (weighted)       & 0.1485 & 0.2191 & \textbf{0.2302} & 0.0000 & 0.0524 \\

\bottomrule
\end{tabular}
\end{adjustbox}
\label{table2}
\end{table*}

%% file: results.tex
The results of the experiments with GPT model can be seen in Table~\ref{table1} and in Table~\ref{table2} for Llama.

This study compares two forecasting paradigms in LLMs: parametric reasoning, where models rely solely on internal knowledge (Experiment 1), and non-parametric reasoning, where they are augmented with retrieved data via RAG (Experiment 2). The results highlight key differences in model capability and integration.

Parametric Forecasting: Internal Knowledge Boundaries
In the parametric setting, GPT-4 outperformed LLaMA across all tasks. GPT-4’s accuracy was particularly high for class-from-fatalities (e.g., 0.83 for Somalia), suggesting it encodes useful general patterns from pretraining. However, categorical class prediction remained weak (macro-F1 < 0.22 across countries), revealing limits in nuanced multi-class reasoning based on internal knowledge alone. LLaMA's parametric performance was consistently low, underscoring its limited capability without external context.

Non-Parametric Forecasting: Benefits of RAG
Adding retrieved context via RAG improved GPT-4’s performance across most tasks and countries. Notable gains were observed in class-from-fatalities (e.g., F1-macro in Israel: 0.16 → 0.33) and binned regression, with reduced MAE in countries like Israel and Iran. These results show GPT-4’s strong ability to integrate structured external inputs.

LLaMA, however, showed minimal or inconsistent gains from RAG. In some cases (e.g., Ethiopia), performance worsened or outputs failed entirely. This suggests that effective non-parametric forecasting requires not just retrieval, but also sufficient model capacity to interpret and incorporate the new information.

Key Takeaways
Parametric forecasting is limited in precision, especially for fine-grained classification.

Non-parametric augmentation improves performance—but only when the model can meaningfully integrate retrieved context.

GPT-4 benefits consistently from non-parametric inputs; LLaMA does not.

%% file: discussion.tex
Our results shed light on the contrasting capabilities of large language models (LLMs) when relying solely on their parametric knowledge (Experiment 1) versus when enhanced with contextual retrieval via a RAG pipeline (Experiment 2). The findings suggest a persistent yet uneven performance gap between these two paradigms, with the degree of improvement varying by model, task formulation, and region.

\textbf{Parametric vs. Contextual Knowledge}
Overall, the integration of non-parametric context via RAG yielded modest but consistent improvements in several tasks, particularly for GPT-4. When comparing classification performance using explicit labels (Experiment 1 – Class Categorical), RAG provided marginal gains or maintained comparable performance for GPT-4. For example, macro F1 scores in Ethiopia and Israel improved (from 0.2091 to 0.2710 and from 0.1471 to 0.2710, respectively), while Sudan and Iran also showed slight improvements. However, in Somalia, performance decreased, indicating that contextual information is not universally beneficial, possibly due to noisy or sparse retrieved content.

The gains were more pronounced in the binned regression task, where RAG substantially improved macro precision and F1 for GPT-4 in regions like Sudan and Israel (e.g., F1 from 0.1464 to 0.3483 and from 0.0000 to 0.4280, respectively), highlighting the value of article-informed context in more granular conflict prediction. Interestingly, RAG did not consistently reduce MAE in fatalities prediction, suggesting that while context can improve classification of conflict intensity, it may not translate directly into better numeric estimation.

LLaMA, in contrast, exhibited much lower baseline performance across all tasks and regions, underscoring its weaker parametric understanding. While RAG improved classification and regression metrics in select cases—such as macro F1 in Somalia and Israel (e.g., F1 macro from 0.0273 to 0.1494 and from 0.2710 to 0.3411, respectively)—the overall performance remained limited. This contrast reaffirms the superior zero-shot reasoning capabilities of GPT-4 and suggests that LLaMA benefits from retrieval mostly in cases where parametric knowledge is clearly insufficient.

\textbf{Model-Specific Trends}
GPT-4 consistently outperformed LLaMA across both experiments and all task formulations. This was especially evident in the categorical classification task, where LLaMA struggled to achieve even modest accuracy scores in regions like Ethiopia and Iran (e.g., 0.0000 and 0.0526 in Exp 2), while GPT-4 remained above 0.25 in all regions. The disparity was also evident in fatalities-derived classes, where GPT-4 maintained strong accuracy (>0.72 in three regions) even in the parametric setting, showing its ability to generalize conflict trends from internalized patterns alone.

Another key difference lies in how the models handle retrieved context. GPT-4 appears to leverage it more reliably, suggesting better instruction-following, contextual grounding, and inference capabilities. LLaMA, while modestly improved with RAG, still struggled with precision and recall, and even regressed in some cases (e.g., binned regression in Israel).

\textbf{Regional Variation vs. Generalization}
Regional differences were pronounced. GPT-4 performed best in Sudan, Iran, and Ethiopia, showing consistent macro F1 improvements across experiments. In contrast, Somalia and Israel were more volatile, possibly due to less consistent article quality, varying event dynamics, or weaker patterns in historical data. This highlights a limitation of both parametric and RAG-enhanced models: context quality and topical consistency are critical.

Interestingly, the performance gap between Experiment 1 and Experiment 2 was not uniform across countries or metrics, indicating that generalization is region-sensitive. For example, in fatalities-based classification, GPT-4 and LLaMA both saw improvements in F1 macro in Sudan and Somalia with RAG, while accuracy and MAE remained stable. This suggests that regional context quality, rather than a global trend, determines the effectiveness of RAG.

%% file: conclusion.tex
This study explores the capacity of large language models (LLMs) to forecast conflict trends and fatalities across multiple regions and timeframes. We compare parametric (zero-shot) forecasting with retrieval-augmented generation (RAG), evaluating two prominent models: GPT-4 and LLaMA-2. Our experiments demonstrate that while both models exhibit some ability to capture broad conflict dynamics, performance improves significantly when external context—structured and unstructured—is incorporated through RAG. This is especially evident in forecasting minority classes such as escalation and de-escalation.

Categorical label prediction remains challenging, however, and numeric fatality estimates are often noisy without careful prompt design and postprocessing. Notably, open-source models like LLaMA-2, when paired with retrieval, can perform competitively with GPT-4 in select contexts—offering promise for use in low-resource or access-constrained environments. Our short-term forecasting setup further suggests that LLMs can be sensitive to recent trends when grounded in localized data.

Overall, this work highlights both the promise and current limitations of LLMs for high-stakes humanitarian forecasting. Future research should explore multilingual retrieval, fine-tuning with domain-specific signals, and human-in-the-loop pipelines to increase reliability and applicability in real-world early warning systems.